\documentclass[preprint,12pt]{elsarticle}
\usepackage{lineno}
\modulolinenumbers[5]  

\usepackage{amssymb}
\usepackage{graphicx}
\usepackage{longtable}
\usepackage{multirow}
\usepackage{subcaption}
\usepackage{amsmath}
\usepackage{url}
\usepackage{hyperref}
\usepackage{adjustbox}
\hypersetup{
     colorlinks   = true,
     citecolor    = blue
}
\usepackage{changepage}
\usepackage{epstopdf}
\usepackage{tabularx}

\usepackage[usenames,dvipsnames,svgnames,table]{xcolor}

\def\@reftitle{}
\newcommand{\reftitle}[1]{\gdef\@reftitle{#1}}

\journal{Applied Sciences}

\graphicspath{{./figs/}}

\begin{document}

\begin{frontmatter}

\title{Learning Eligibility in Cancer Clinical Trials \\ using Deep Neural Networks}
\author{Aurelia Bustos}
\ead{aurelia@medbravo.org}
\author{Antonio Pertusa\corref{cor1}}
\ead{pertusa@dlsi.ua.es}
\address{Pattern Recognition and Artificial Intelligence Group (GRFIA), Department of Software and Computing Systems, University Institute for Computing Research, University of Alicante, E-03690 Alicante, Spain}
\cortext[cor1]{Corresponding author.}

\begin{abstract}
Interventional cancer clinical trials are generally too restrictive, and some patients are often excluded on the basis of comorbidity, past or concomitant treatments, or the fact that they are over a certain age. The efficacy and safety of new treatments for patients with these characteristics are, therefore, not defined. In this work, we built a model to automatically predict whether short clinical statements were considered inclusion or exclusion criteria. We used protocols from cancer clinical trials that were available in public registries from the last 18 years to train word-embeddings, and we constructed a~dataset of 6M short free-texts labeled as eligible or not eligible. A text classifier was trained using deep neural networks, with pre-trained word-embeddings as inputs, to predict whether or not short free-text statements describing clinical information were considered eligible. We additionally analyzed the semantic reasoning of the word-embedding representations obtained and were able to identify equivalent treatments for a type of tumor analogous with the drugs used to treat other tumors. We show that representation learning using {deep} neural networks can be successfully leveraged to extract the medical knowledge from clinical trial protocols for potentially assisting practitioners when prescribing treatments.
\end{abstract}


\end{frontmatter}

\section{Introduction}
Clinical trials (CTs) provide the evidence needed to determine the safety and effectiveness of new medical treatments. These trials are the bases employed for clinical practice guidelines \cite{nccn_2017} and greatly assist clinicians in their daily practice when making decisions regarding treatment. However, the eligibility criteria used in oncology trials are too restrictive \cite{Jin_2017}. Patients are often excluded on the basis of comorbidity, past or concomitant treatments, or the fact they are over a certain age, and~those patients that are selected do not, therefore, mimic clinical practice. This signifies that the results obtained in CTs cannot be extrapolated to patients if their clinical profiles were excluded from the clinical trial protocols. Given the clinical characteristics of particular patients, their type of cancer, and the intended treatment, discovering whether or not they are represented in the corpus of CTs that is available requires the manual review of numerous eligibility criteria, which is impracticable for clinicians on a daily basis. 

The process would, therefore, greatly benefit from an evidence-based clinical decision support system (CDSS). 
{Briefly, a CDSS could scan free-text clinical statements from medical records and output the eligibility of the patient in both completed or ongoing clinical trials based on conditions, cancer molecular subtypes, medical history, and treatments. Such a CDSS would have the potential advantages of (1) assessing the representation of the patient's case in completed studies to more confidently extrapolate study results to each patient when prescribing a treatment in clinical practice, and (2) screening a patient's eligibility for ongoing clinical trials.}

In this work, we constructed a dataset using the clinical trial protocols published in the largest public registry available, and used it to train and validate a model that is able to predict whether short free-text statements (describing clinical information, like patients’ medical history, concomitant~medication, type and features of tumor, such as molecular profiles, cancer therapy, etc.) are considered as \textit{Eligible} or \textit{Not Eligible} criteria in these trials. This model is intended to inform clinicians whether the results obtained in the CTs---and, therefore, the recommendation in the standard guidelines---can be confidently applied to a particular patient. The ultimate goal of this work is to assess whether representation learning using {deep} neural networks could be successfully applied to extract the medical knowledge available on clinical trial protocols, thus paving the way toward more involved and complex projects.


In the present work, the text was first preprocessed in order to construct training and validation sets. {After extracting bigrams and word-embeddings (which are commonly used techniques used to generate semantic representations), we explored different state-of-the-art classification methods (FastText, Convolutional Neural Networks (CNN), Support Vector Machines (SVM), and  k-Nearest Neighbors (kNN))}. Finally, after validating and comparing the final classifiers, the model was further tested against an independent testing set. 

{
The main contributions of this work are as follows:
\begin{itemize}
\item We propose a method to learn the eligibility for cancer clinical trials collected in last 18 years.
\item Several classifiers (FastText, CNN, SVM, and kNN) are evaluated using word-embeddings for eligibility classification.
\item Using learned deep representations, CNN and kNN (in this case, with average word-embeddings) obtain a similar accuracy, outperforming the other methods evaluated.
\item Representation learning extracts medical knowledge in cancer clinical trials, and~word-embeddings are suitable to detect tumor type and treatment analogies. 
\item In addition, word-embeddings are also able to cluster semantically related medical concepts.
\end{itemize}
}

The remainder of this paper is organized as follows. Section \ref{sec:related} provides a brief review of the methods related to the proposed work. Section \ref{sec:materialsmethods} describes the dataset constructed and the methodology used, including the details employed to train the embeddings and the text classifiers. The evaluation results are detailed in Section \ref{sec:results}, along with an analysis of the word-embeddings that were learned.  Finally, Section \ref{sec:conclusions} addresses our conclusions and future work. 

\section{Related Work}
\label{sec:related}

Artificial intelligence methods {include, among others,} rule-based systems, traditional machine learning algorithms, and representation learning methods, {such as deep learning architectures.}

Rule-based approaches in Natural Language Processing (NLP) seek to {encode} biomedical knowledge in formal languages in such a way that a computer can automatically reason about text statements in these formal languages using logical inference rules. {MetaMap \cite{aronson2001effective} is a widely known rule-based processing tool in the broader domain of biomedical language. It is} a named-entity recognition system which identifies concepts from the Unified Medical Language System Metathesaurus in text (and MetaMap Lite \cite{demner2017metamap}), the clinical Text Analysis and Knowledge Extraction System (cTAKES \cite{savova2010mayo}), and DNorm \cite{leaman2013dnorm}. Many systems have been built upon those tools. For example, in~\cite{savova_2017_deepphe}, an NLP System for Extracting Cancer Phenotypes from Clinical Records was built to describe cancer cases on the basis of a mention-annotation pipeline based on an ontology and a cTAKES system, and a phenotype summarization pipeline based on the Apache Unstructured Information Management Architecture (UIMA \cite{mcewan2016nlp}).

With regard to the specific domain of clinical trials, prior work has focused on the problem of formalizing eligibility criteria using rule-based approaches and obtaining a computational model that could be used for clinical trial matching and other semantic reasoning tasks. Several languages could be applied in order to express eligibility criteria, such as Arden syntax, Gello, and ERGO, among others. Weng et al. \cite{Weng_2010} presented a rich overview of existing options. SemanticCT allows the formalization of eligibility criteria using Prolog rules \cite{huang2013semanticct}. Milian et al. \cite{milian_2015_enhancing} applied ontologies and regular expressions to express eligibility criteria as semantic queries. However, the problem of structuring eligibility criteria in clinical trials so as to obtain a generalizable model still remains unsolved.

Devising formal rules and representations with sufficient complexity to accurately describe biomedical knowledge is problematic. As an example, the problem with discrete representations in biomedical taxonomies and ontologies is that they miss nuances and new words (e.g., it is impossible for them to keep up to date with the new drugs in cancer research). In addition, they are subjective, require human labor to create and adapt them, and it is hard to compute word similarity accurately. In~order to solve these issues, machine learning methods can be trained to acquire this knowledge by extracting patterns from raw data.

In traditional machine learning, the features employed to train algorithms, such as SVMs or kNN, are {usually given}, while in representation learning (deep learning methods such as CNN), these features are learned~\cite{Goodfellow2016DeepLearning}. Nonetheless, many factors regarding variation influence the semantic interpretation of the biomedical language, thus making it very difficult to extract high-level abstract features from raw text. Deep learning solves this central problem by means of representation learning by introducing representations that are expressed in terms of other simpler representations.

Deep learning models are beginning to achieve greater accuracy and semantic capabilities \cite{naturedl} than the prior state of the art with regard to various biomedical tasks, such as automatic clinical text annotation and classification. For example, a recent work  \cite{mullenbach2018explainable} presented an attentional convolutional network that predicts medical codes from clinical text. It aggregates information from throughout the document using a CNN, and then uses an attention mechanism to select the most relevant segments for each of the thousands of possible codes. With regard to clinical text classification tasks, \cite{hughes2017medical} proposed an approach with which to automatically classify a clinical text at a sentence level using deep CNNs to represent complex features. 

To the best of our knowledge, this work is the first reported study to explore the use of deep learning techniques in order to directly achieve a semantic interpretation of eligibility criteria in clinical trials.  In contrast to classic NLP approaches, to build the model, we omitted the constraints and limitations of previous steps, such as tokenization, stemming, syntactic analysis, named entity recognition (NER), the tagging of concepts to ontologies, rule definition, or the manual selection of features.

\section{Materials and Methods}
\label{sec:materialsmethods}

\begin{figure}
\centering
\includegraphics[width=0.95\linewidth]{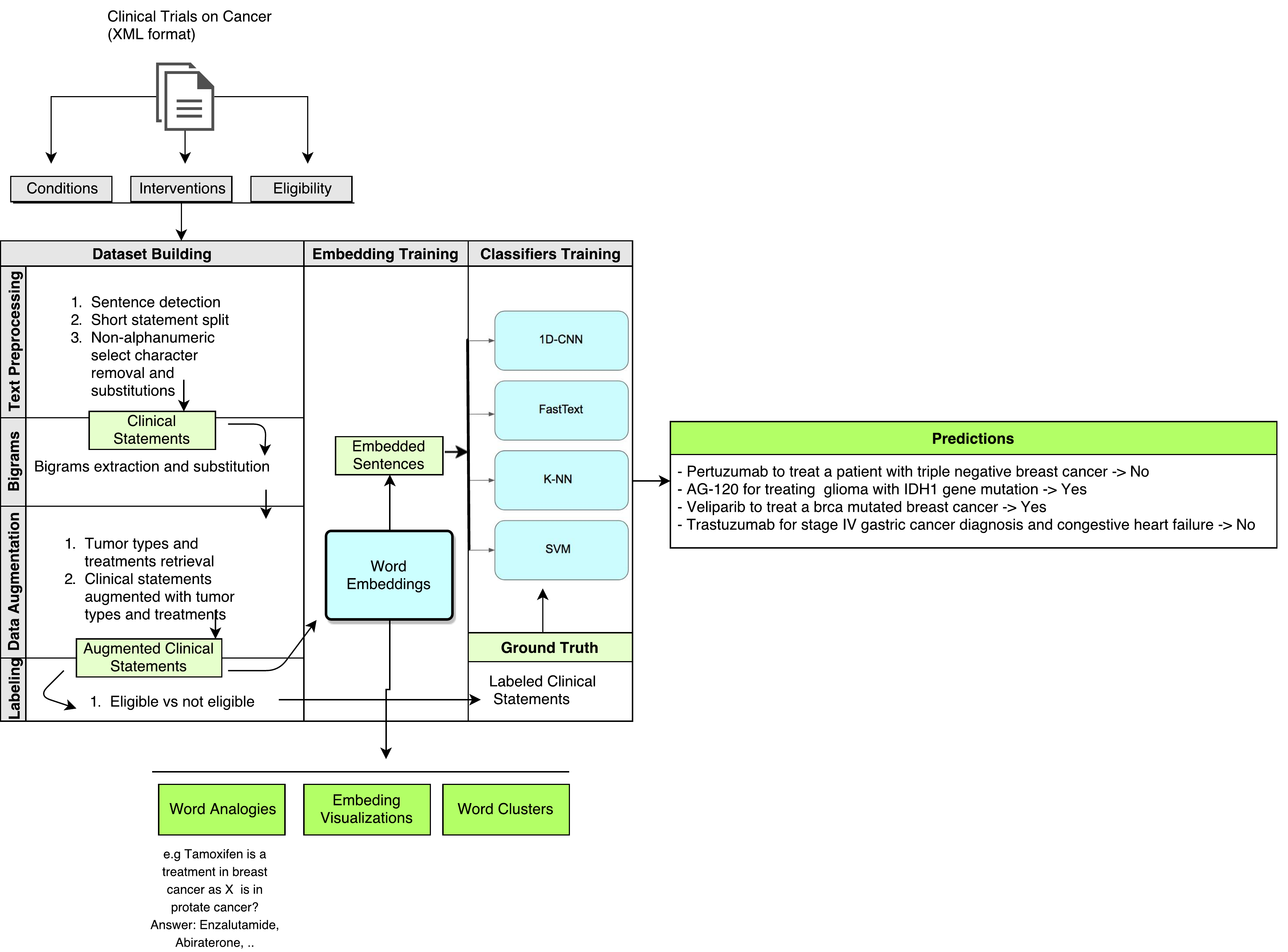}
\caption{System Architecture: The objective of the final model is to predict whether or not short clinical statements concerning the type of tumor, including the molecular profile, the oncologic treatment, the~medical history, or concomitant medication, were included in clinical trials.}
\label{figCapstone} 
\end{figure}

{The system architecture is shown in Figure \ref{figCapstone}. Clinical trials statements were first preprocessed as described in Section \ref{sec:datasetbuilding}. Then, word-embeddings were trained,
as shown in Section \ref{sec:embed_train}, and classification to obtain the eligibility prediction is detailed in Section \ref{sec:class_train}.  }

\subsection{Dataset Building}
\label{sec:datasetbuilding}

A total of 6,186,572 labeled clinical statements were extracted from 49,201 interventional CT protocols on cancer (the URL for downloading this dataset is freely available at \url{https://clinicaltrials.gov/ct2/results?term=neoplasm&type=Intr&show_dow}). 
Each CT downloaded is an XML file that follows a structure of fields defined by an XML schema of clinical trials \cite{CT_XML_2017}. The relevant data for this project are derived from the intervention, condition, and eligibility fields written in unstructured free-text language. The~information in the eligibility criteria---both exclusion and inclusion criteria---are sets of phrases and/or sentences displayed in a free format, such as  paragraphs, bulleted lists, enumeration lists, etc. None of these fields use common standards, nor do they enforce the use of standardized terms from medical dictionaries and ontologies. Moreover, the language had the problems of both polysemy and~synonymy. 

\begin{figure}
\centering
\includegraphics[width=\linewidth]{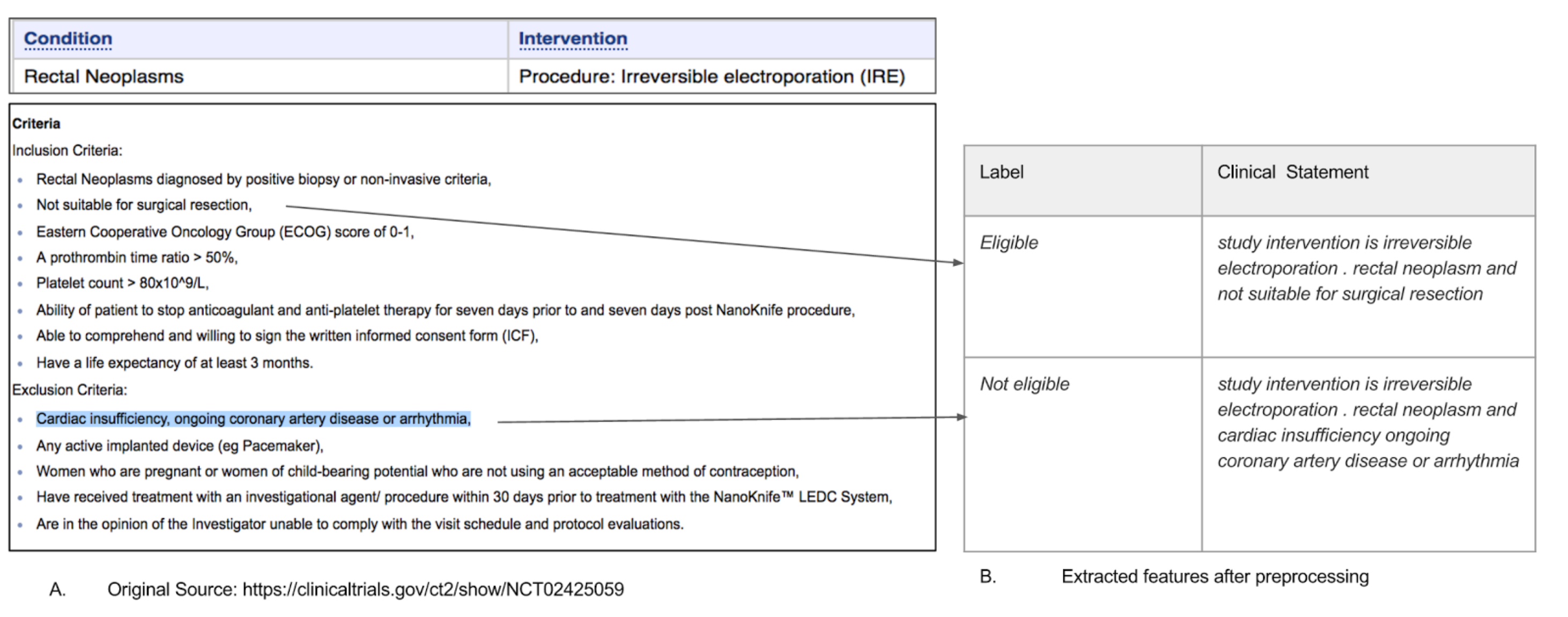}
\caption{Extraction of labeled short clinical statements. The two {example} criteria indicated with the arrows were extracted from their original source, preprocessed, and labeled.}
\label{fig2} 
\end{figure}

The original data were exploited by merging eligibility criteria together with the study condition and intervention, and subsequently transforming them into lists of short labeled clinical statements {that consisted of two extracted features (see example in Figure \ref{fig2}), the label (Eligible or Not Eligible), and the processed text that included the original eligibility criterion merged with the study interventions and the study conditions}. These processes are detailed in the following section.

\subsubsection{Text Preprocessing}
\label{sec:preprocessing}

We transformed all the eligibility criteria into sequences of plain words (and bigrams) separated by a whitespace. Each eligibility criterion was augmented with information concerning study intervention and cancer type, {as illustrated in Figure \ref{fig2}}. This was done by:

\begin{itemize}
  \item Splitting text into statements: The implementation took into consideration different kinds of bullets and lists, and not mistakenly splitting into sentences common abbreviations used in mutations and other medical notations, which include dots, semicolons, or hyphens.
  \item Removing punctuation, whitespace characters, all non-alphanumeric symbols, separators, and single-character words from the extracted text. All the words were lowercase. We decided not to remove stop words because many of them, such as ``or”, ``and”, ``on”, were semantically relevant to the clinical statements. 
   \item Transforming numbers, arithmetic signs (+/$-$), and comparators (>, <, =, ...) into text. 
\end{itemize}

{
In order to filter out nonrelevant or useless samples, we discarded all the studies where the conditions did not include any of the tokens or suffixes in ``cancer”, ``neoplasm”, ``oma”, or ``tumor”. Given that the presence or absence of redundancy in eligibility criteria, both intra- or interstudy, is relevant information to be learned by the model, we did not filter out samples by this criteria, so that the original redundancy distribution was preserved in the dataset. 
Because preprocessing the entire dataset is a costly process, for those readers interested in reproducing this work but would like to skip the preprocessing steps, we made publicly available a random preprocessed subsample (\url{https://www.kaggle.com/auriml/eligibilityforcancerclinicaltrials}, at section Data, Download all) 
of $10^6$ samples  
(for details, see Section \ref{sec:preprocessing}). 
}

\subsubsection{Bigrams}

In the scope of this work, we define bigrams as commonly found phrases that are very frequent in medicine. Some frequent bigrams were detected and replaced in the text. Bigrams can represent idiomatic phrases (frequently co-occurring tokens) that are not compositions of the individual words. Feeding them as a single entity to the word-embedding rather than each of its word separately, therefore, allows these phrase representations to be learnt. In our corpus, excluding common terms, such as stop words, was unnecessary when generating bigrams. Some examples of bigrams in this dataset are: sunitinib malate, glioblastoma multiforme, immuno histochemistry, von willebrand, dihydropyrimidine dehydrogenase, li fraumeni, etc.

Phrase (collocation) detection was carried out using the GenSim API \cite{rehurek_lrec}. The threshold parameter defines which phrases will be detected on the basis of their score. The score formula applied \cite{mikolov_2013_distributed} is:

\begin{equation}
 score(w_{i},w_{j})=\frac{count(w_{i},w_{j}) - \delta}{count(w_{i}) \cdot count(w_{j})}
\end{equation}

For this dataset, after several tests, the most suitable threshold was set to 500, and the discounting coefficient $\delta$ was based on a min count of 20. The discounting factor prevents the occurrence of too many phrases consisting of very infrequent words. A total of 875 different bigrams were retrieved from the corpus and substituted in the text.

\subsubsection{Data Augmentation}

In this work, data augmentation consisted of adding the cancer types and interventions being studied to each criterion using statements such as: ``patients diagnosed with [cancer type]''. 

In the case of CTs that studied multiple cancer types or interventions, we replicated each criterion for each intervention and condition, increasing the number of prototypes.

\subsubsection{Labeling}

{After preprocessing and cleaning the data, the available set had 6,186,572 short clinical statements containing a total of 148,038,397 words. The vocabulary consisted of 49,222 different words. {Each~statement had in average 23.9 words with a range from 6 to 439 words. The distribution of number of words by statement had a mean = 23.9, variance = 171.3, skewness = 3.13, and kurtosis = 21.05. }}

For the ground truth, we automatically labeled the clinical statements - previously processed from the eligibility criteria, study conditions, and interventions - as ``Eligible” (inclusion criterion) or ``Not Eligible” (exclusion criterion) on the basis of:

\begin{itemize}
\item Their position in relation to the sentences ``inclusion criteria'' or ``exclusion criteria'', which~usually preceded the respective lists. If those phrases were not found, then the statement was labeled~``Eligible''.
\item Negation identification and transformation: negated inclusion criteria starting with ``no” were transformed into positive statements and labeled ``Not Eligible''. All other possible means of negating statements were expected to be handled intrinsically by the classifier.
\end{itemize} 

The classes were unbalanced, and only 39\% of them were labeled as Not Eligible, while 61\% were labeled as Eligible.  As the dataset was sufficiently large, we used random balanced undersampling~\cite{LingS_2007_Cost} to correct it, resulting in a reduced dataset with 4,071,474 labeled samples. The “eligibility” variable containing the text for each criterion, as expected in NLP, has a highly sparse distribution and only 450~entries were repeated.



\subsection{Embedding Training}
\label{sec:embed_train}

We used two different approaches (FastText \cite{Bojanowski_2016_Enriching} and Gensim \cite{rehurek_lrec}) to generate Word2Vec embeddings based on the skip-gram and CBOW models \cite{Mikolov13}. Word2vec \cite{mikolov_2013_distributed} is a predictive model that uses raw text as input and learns a word by predicting its surrounding context (continuous BoW model) or predicts a word given its surrounding context (skip-gram model) using gradient descent with randomly initialized vectors. In this work, we used the Word2Vec skip-gram model.
The main differentiating characteristic of FastText embeddings, which apply char $n$-grams, is that they take into account the internal structure or words while learning word representations \cite{Bojanowski_2016_Enriching}. This~is especially useful for morphologically rich languages. FastText models with char $n$-grams perform significantly better when carrying out syntactic tasks than semantic tasks, because the syntactic questions are related to the morphology of the words. 

We explored different visualizations projecting the trained word-embeddings into the vector space (Sections \ref{sec:tsnesubset} and \ref{sec:interactivevis}), grouped terms in semantic clusters (Section \ref{sec:clusters}), and qualitatively evaluated the embeddings according to their capacity to extract word analogies (Section \ref{sec:wordanalogies}).

Table \ref{tab:w2v-hyper} shows the best hyperparameters found to generate 100 dimensional embeddings with the FastText and Gensim Word2Vec models. {A random search strategy \cite{bergstra2012random} was used in order to optimize the values of these parameters.} The Gensim model was trained with three workers on a final vocabulary of 22,489 words using both skip-grams and CBOW models. 

\begin{table}
\centering
\caption{Word2Vec hyperparameters using FastText and GenSim. {Optimization was performed using random search \cite{bergstra2012random}.}}
\begin{footnotesize}
\begin{tabular}{lll}
\hline
\textbf{Hyperparameter }					 & \textbf{FastText} & \textbf{GenSim}  \\
\hline
Learning rate                         & 0.025   & 0.025      \\
Size of word vectors                  & 100     & 100      \\
Size of the context window            & 5       & 5      \\
Number of epochs                      & 5       & 5      \\
Min. number of word occurrences        & 5       & 5      \\
Num. of negative sampled              & 5       & 5      \\
Loss function                         & negative sampling & negative sampling 
		\\
Sampling threshold                    & $10^{-4}$  & $10^{-3}$ \\
Number of buckets                     & 2,000,000 &    \\
Minimum length of char $n$-gram             & 3         &    \\
Maximum length of char $n$-gram             & 6         &    \\
Rate of updates for the learning rate & 100 & \\      
\hline
\end{tabular}
\end{footnotesize}
\label{tab:w2v-hyper}
\end{table}

\subsection{Classifier Training}
\label{sec:class_train}
{
Once the word-embeddings were extracted, the next stage consisted of sentence classification. For this, we explored four methods: Deep Convolutional Neural Networks \cite{naturedl} with or without pre-trained word-embeddings at the input layer, FastText \cite{Joulin_2016_Bag}, Support Vector Machines (SVM), and k-Nearest Neighbors (kNN).}

Learning curves were built for {all} models with increasing dataset sizes (1K, 10K, 100K, 1M, and 4.07 
M samples). Each dataset was {sampled from the full dataset, applying random balanced sampling so that, for each resulting dataset, both label classes (``Eligible'' and ``Not Eligible'') had the same number of samples}. We split each dataset into 80\% samples for the training set and 20\% for the test set. A standard 5-fold cross-validation was then performed for each dataset size.

Because the accuracy concerning sentence classification depends on the dataset evaluated and we were unable to find any previous reports that used the present corpus for text classification, there are no clearly defined benchmarks with which to perform a comparison. 

For example, in different domains, the reported accuracy for classifying the ``Hacker News'' posts into 20 different categories using a similar method was 95\%, while in the case of “Movie reviews”, the reported performance was 81.5\% \cite{Kim_14_Convolutional}. 
In the medical domain, a high-performance model is potentially useful in a CDSS. Using previously published computer-aid systems and related work \cite{zhang_2017_automated,ni_2015_increasing,Das_2017_Using} as a basis, we defined the minimum target as an accuracy of  90\%, and a Cohen's Kappa with a minimum of [0.61--0.80] for substantial agreement, or [0.81--1] for an almost perfect agreement~\cite{landis_1977_measurement}.

\subsubsection{FastText}

FastText \cite{Bojanowski_2016_Enriching,Joulin_2016_Bag} for supervised learning is a computationally efficient method that starts with an~embedding layer which maps the vocabulary indexes into $d$ dimensions or, alternatively, it can use pre-trained word vectors. It then adds a global average pooling layer, which averages the embeddings of all the words in the sentence. Finally, it projects it onto a single unit output layer and squashes it with a sigmoid. 

\subsubsection{Convolutional Neural Network}

In the first experiment, the pre-trained word-embeddings were used as the input for the 1D CNN model, which has a final dense output layer. In a different experiment, we also trained the word-embeddings for our classification task from scratch. As the training data was sufficiently large and the vocabulary coverage was also appropriate for the cancer research domain, it was expected that the model would benefit from training the embeddings in this particular domain. 

We used the Keras \cite{chollet_2015_keras} library to build a CNN topology (see Table \ref{tab:cnn}), inspired by the text classifier model for the 20 Newsgroup datasets \cite{keras_2017_examples}. After the necessary adaptations, we followed the steps shown below:



\begin{enumerate}
\item Convert all the sentences in the dataset into sequences of word indexes. A \textit{word index} is simply an integer identifier for the word. We considered only the top 20,000 most commonly occurring words in the dataset, and truncated the sequences to a maximum length of 1000 words.
\item Shuffle, stratify, and split sequences of word indexes into training (80\%) and validation sets (20\%).
\item Prepare an \textit{embedding matrix} which contains at index $i$ the embedding vector for the word from index $i$. We loaded this embedding matrix into an embedding layer which was frozen (i.e., its weights, the embedding vectors, were not updated during training).
\item A 1D CNN ending in a Softmax layer with two classes was built on top of it.
\item During training, the data were shuffled with random seed before each epoch (we used 10 epochs).
\end{enumerate}

\begin{table}
\centering
\caption{CNN topology used in this work. {The architecture was chosen after evaluating the accuracy on the test set using different kernel sizes, number of layers, activation functions, etc.}}
\begin{footnotesize}
\begin{tabular}{ll}
\hline
\textbf{Layer} & \textbf{Description} \\
\hline
Input & 1000 $\times$ 100 dimensional embedded word sequences \\
Convolution & 128  5 $\times$ 1 convolutions with stride 1 and ReLu activation        \\
Max Pooling &5 $\times$ 1 max pooling with stride 1 \\
Convolution & 128  5 $\times$ 1 convolutions with stride 1 and ReLu activation        \\
Max Pooling &5 $\times$ 1 max pooling with stride 1 \\
Convolution & 128  5 $\times$ 1 convolutions with stride 1 and ReLu activation        \\
Max Pooling &35 $\times$ 1 max pooling with stride 1 \\
Fully Connected & 128 fully connected layer with ReLu activation \\
Fully Connected & 2 fully connected layer with Softmax activation \\
\hline
\end{tabular}
\end{footnotesize}
\label{tab:cnn}
\end{table}

\subsubsection{SVM}
{
A support vector machine \cite{cortes1995support} constructs a hyperplane or set of hyperplanes in a high- or infinite-dimensional space, which can be used for classification. Intuitively, a good separation is achieved by the hyperplane that has the largest distance to the nearest training data points of any class (so-called functional margin), since, in general, the larger the margin, the lower the generalization error of the classifier. We trained an SVM model with the following hyper-parameters selected using exhaustive grid-search optimization: penalty parameter C or the error term = 1, kernel = rbf, kernel gamma coefficient = 1,  shrinking heuristic = True, tolerance for stopping criterion = 0.001.}

{
For each short clinical statement, its pre-trained word-embeddings (obtained with FastText using the skip-gram model, as explained in Section \ref{sec:embed_train}) 
were used to calculate an average vector of dimension 100 for each clinical statement. Therefore, given a statement, an average vector of word-embeddings serves as input to the SVM. This representation was chosen to reduce the dimensionality of the input~data. }

\subsubsection{kNN}
{Neighbors-based classification is a type of instance-based learning or nongeneralizing learning: it~does not attempt to construct a general internal model, but simply stores instances of the training data. Classification is computed from a simple majority vote of the nearest neighbors of each point: a~query point is assigned the data class which has the most representatives within the nearest neighbors of the point.
}
{The same input data used for SVM were evaluated using kNN. We trained a kNN model with the following hyper-parameters selected using exhaustive grid-search optimization: number of neighbors = 3, uniform weight for all points in each neighborhood, and Euclidean distance metric.}

\section{Results}
\label{sec:results}
\vspace{-6pt}
\subsection{Metrics}

The performance of the models was calculated using the F-measure ($F_1$), precision and recall, the~confusion matrix, and the coefficient of agreement. 

Precision (also called positive predictive value) is the fraction of retrieved instances that are relevant. Recall (also known as sensitivity) is the fraction of relevant instances that are retrieved. Precision, Recall, and $F_1$, which is the harmonic mean of precision and sensitivity, are calculated as:

\begin{align*}
   \text{Precision} &= \frac{\textrm{TP}}{\textrm{TP} + \textrm{FP}} \\
   \text{Recall}    &= \frac{\textrm{TP}}{\textrm{TP} + \textrm{FN}} \\
   F_1 & = 2 \times \frac{ \text{Precision} \cdot \text{Recall}} { \text{Precision} + \text{Recall}} 
\end{align*}



\noindent where TP (true positives) denotes the number of correct predictions, FP (false positives) is the number of ``Not Eligible” labels wrongly predicted as ``Eligible”, and FN (false negatives) is the number of ``Eligible” labels wrongly declared as ``Not Eligible”. 

Cohen's Kappa ($\kappa$) is a statistic that measures the inter-rater agreement for qualitative (categorical) items. It is generally considered to be a more robust measure than a simple percent agreement calculation, since $\kappa$ takes into account the possibility of the agreement occurring by chance \cite{mchugh_2012_interrater}. It is calculated as:

\begin{equation}
\kappa = \frac{p_{o} - p_{e}}{1 - p_{e}} = 1 -\frac{1 - p_{o}}{1 - p_{e}}
\end{equation}

\noindent where $p_o$ is the relative observed agreement among raters, and $p_e$ is the hypothetical probability of a~chance agreement, using the observed data to calculate the probabilities of each observer randomly yielding each category. 

\subsection{Model Evaluation and Validation}

\begin{figure}
\begin{subfigure}{.5\textwidth}
  \centering
  \includegraphics[width=.85\linewidth]{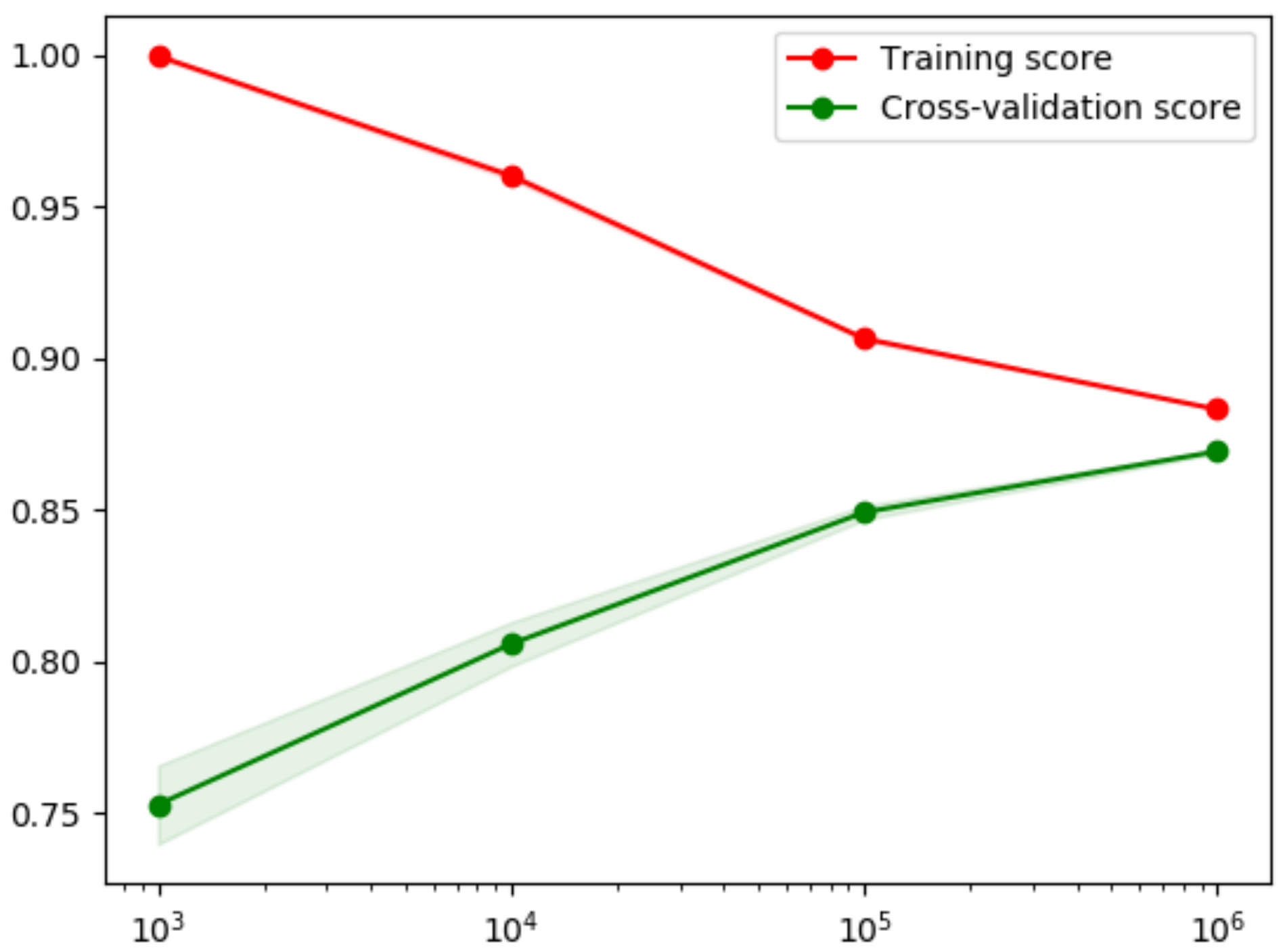}
  \caption{FastText}
  \label{fig4}
\end{subfigure}%
\begin{subfigure}{.5\textwidth}
  \centering
  \includegraphics[width=.85\linewidth]{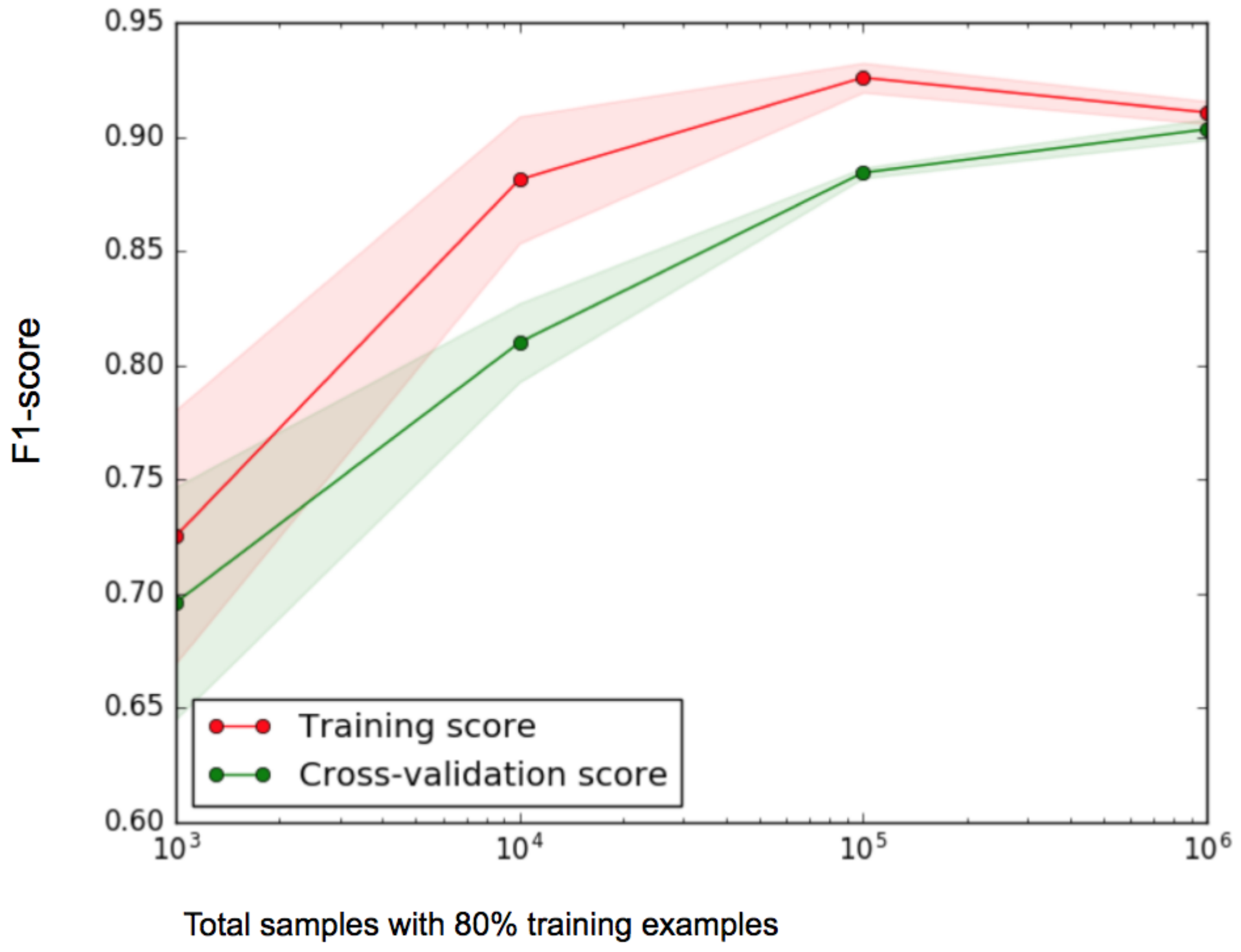}
  \caption{CNN}
  \label{fig5}
\end{subfigure}
\begin{subfigure}{.5\textwidth}
  \centering
  \includegraphics[width=.85\linewidth]{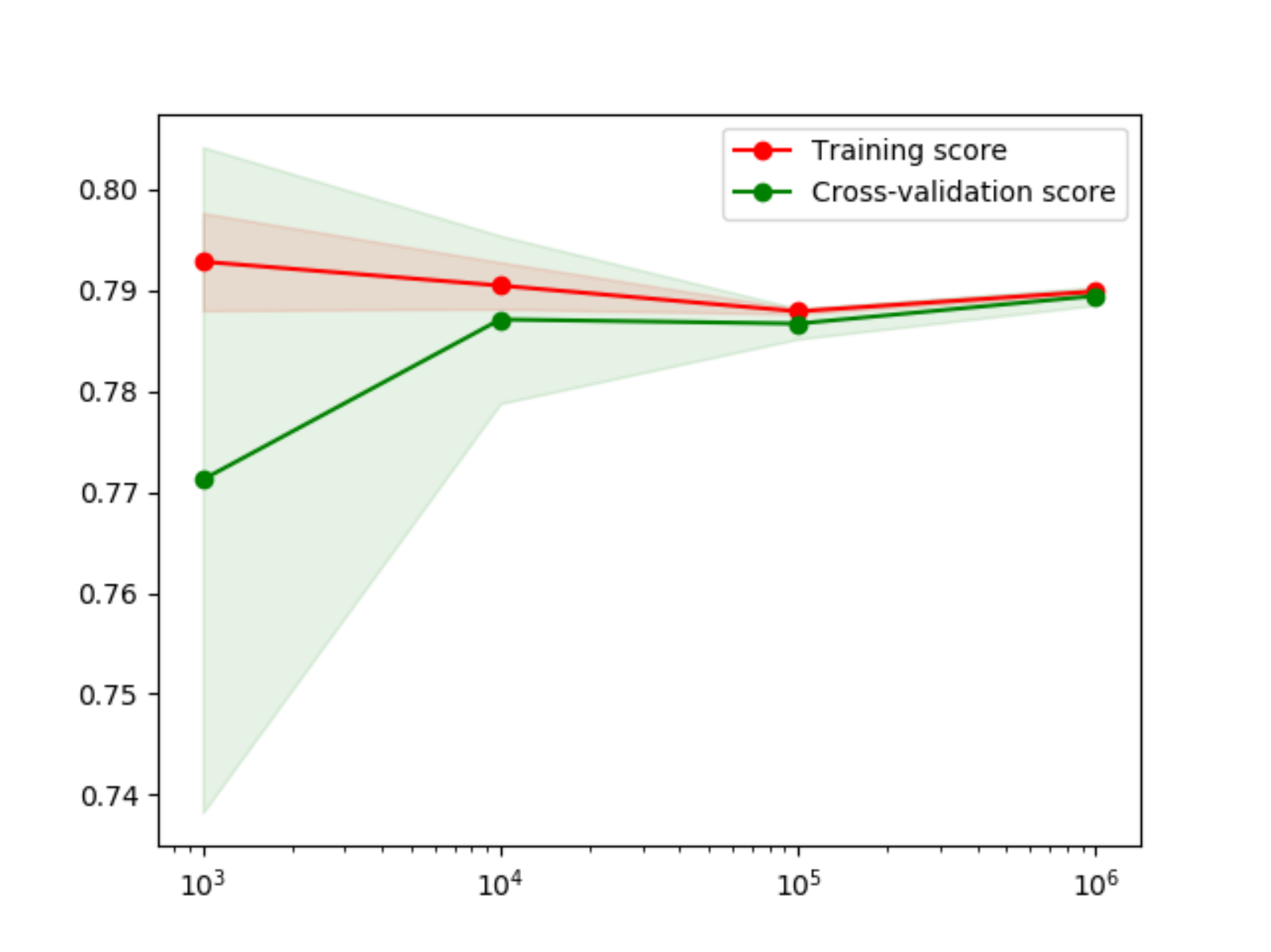}
  \caption{SVM}
  \label{SVM_LC}
\end{subfigure}
\begin{subfigure}{.5\textwidth}
  \centering
  \includegraphics[width=.85\linewidth]{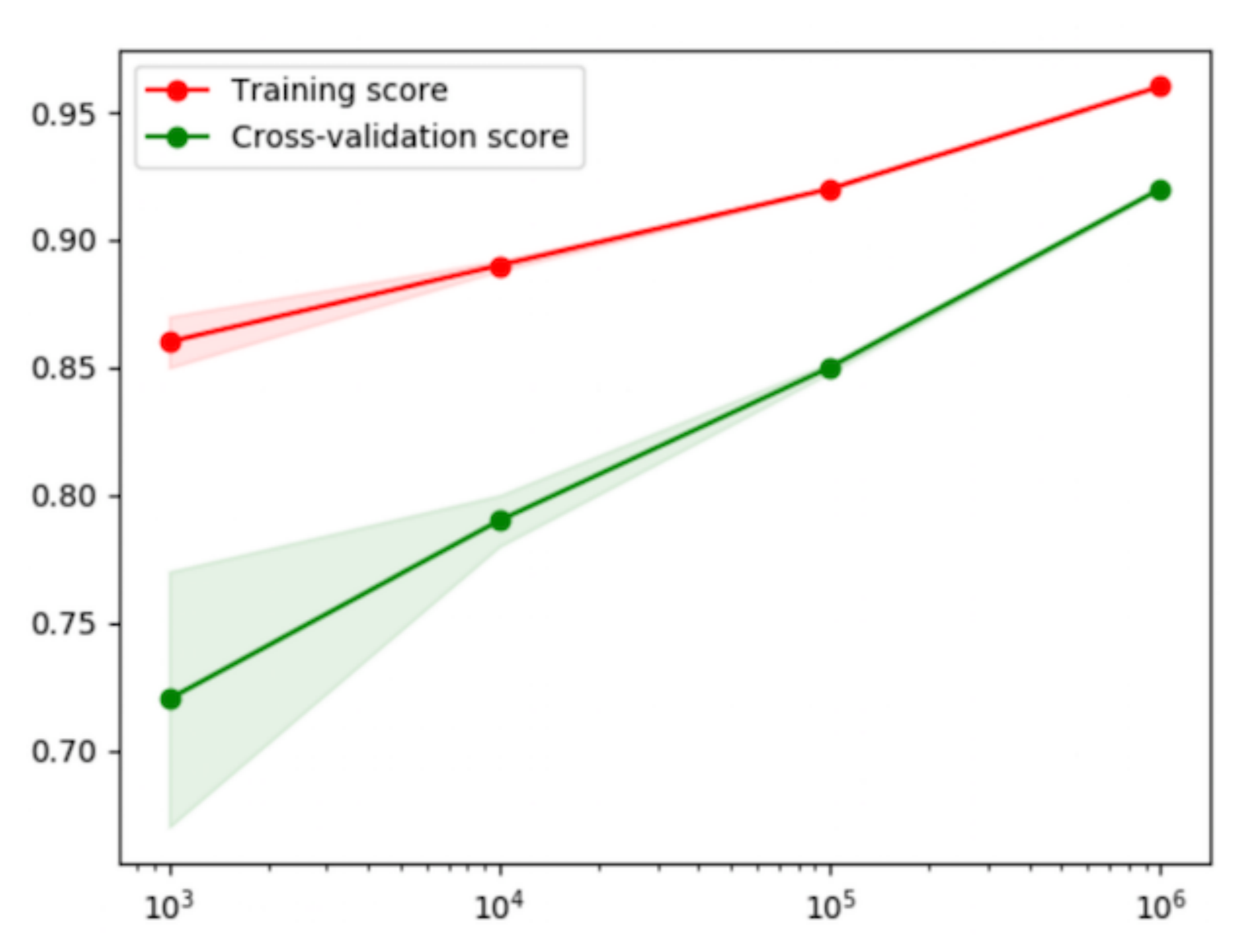}
  \caption{kNN}
  \label{kNN_LC}
\end{subfigure}
\vspace{-16pt}
\caption{{Learning curves with 5-fold cross-validation on the FastText, Convolutional Neural Network (CNN), Support Vector Machine (SVM), and k-Nearest Neighbors (kNN) classifiers. Horizontal axes show the total training samples, whereas vertical axes are the $F_1$-score. The green region represents the standard deviation of the models.}}
\label{fig:learningcurves}
\end{figure}

{All the models were evaluated using different configurations of hyper-parameters, and the best results obtained for each classifier are given in Table \ref{table_results}. This section details the classifier settings to get these results, and analyzes the learning curves that can be seen in Figure \ref{fig:learningcurves}.}

\begin{table}
\centering
\caption{{Overall results on the validation set for all the classifiers using a dataset of  $10^6$ samples and the full dataset (4.1 $\times$ $10^6$) samples. Both experiments were performed using 20\% of the prototypes for validation and 80\% for training. The best results are marked in bold.}}
\label{table_results}
\begin{footnotesize}
\begin{tabular}{cccccc}
\hline
\textbf{Classifier} & \textbf{Dataset Size}  & \textbf{Precision} & \textbf{Recall} & \boldmath{$F_1$} & \textbf{Cohen's} \boldmath{$\kappa$} \\
\hline
\multirow{2}{*}{FastText} & $10^6$     & 0.88               & 0.86            & 0.87    &         0.75      \\
& 4.1 $\times$ $10^6$ & 0.89 & 0.87 & 0.88 & 0.76  \\
\hline

\multirow{2}{*}{CNN} & $10^6$ & 0.88 & 0.88 & 0.88 & 0.76 \\
 & 4.1 $\times$ $10^6$ & 0.91 & 0.91 & 0.91 & 0.83 \\
\hline

\multirow{2}{*}{SVM} & $10^6$ & 0.79 & 0.79 & 0.79 & 0.57 \\
& 4.1 $\times$ $10^6$ & 0.79 & 0.79 & 0.79 & 0.58 \\
\hline

\multirow{2}{*}{kNN} & $10^6$ & 0.92 & 0.92 & 0.92 & 0.83 \\
& 4.1 $\times$ $10^6$ & 0.93 & 0.93 & \textbf{0.93} & \textbf{0.84} \\
\hline
\end{tabular}
\end{footnotesize}
\end{table}
\unskip


\subsubsection{FastText Classifier Results}

In order to choose the parameters for the FastText model, we compared the $F_1$ between successive experiments. 
Using a random search \cite{bergstra2012random} strategy for hyperparameter search, the best results were obtained with 100 dimensions and a learning rate of 0.1, as shown in Table \ref{tab:w2v-hyper-classifier}. 

\begin{figure}
\begin{subfigure}{.5\textwidth}
  \centering
  \includegraphics[width=.85\linewidth]{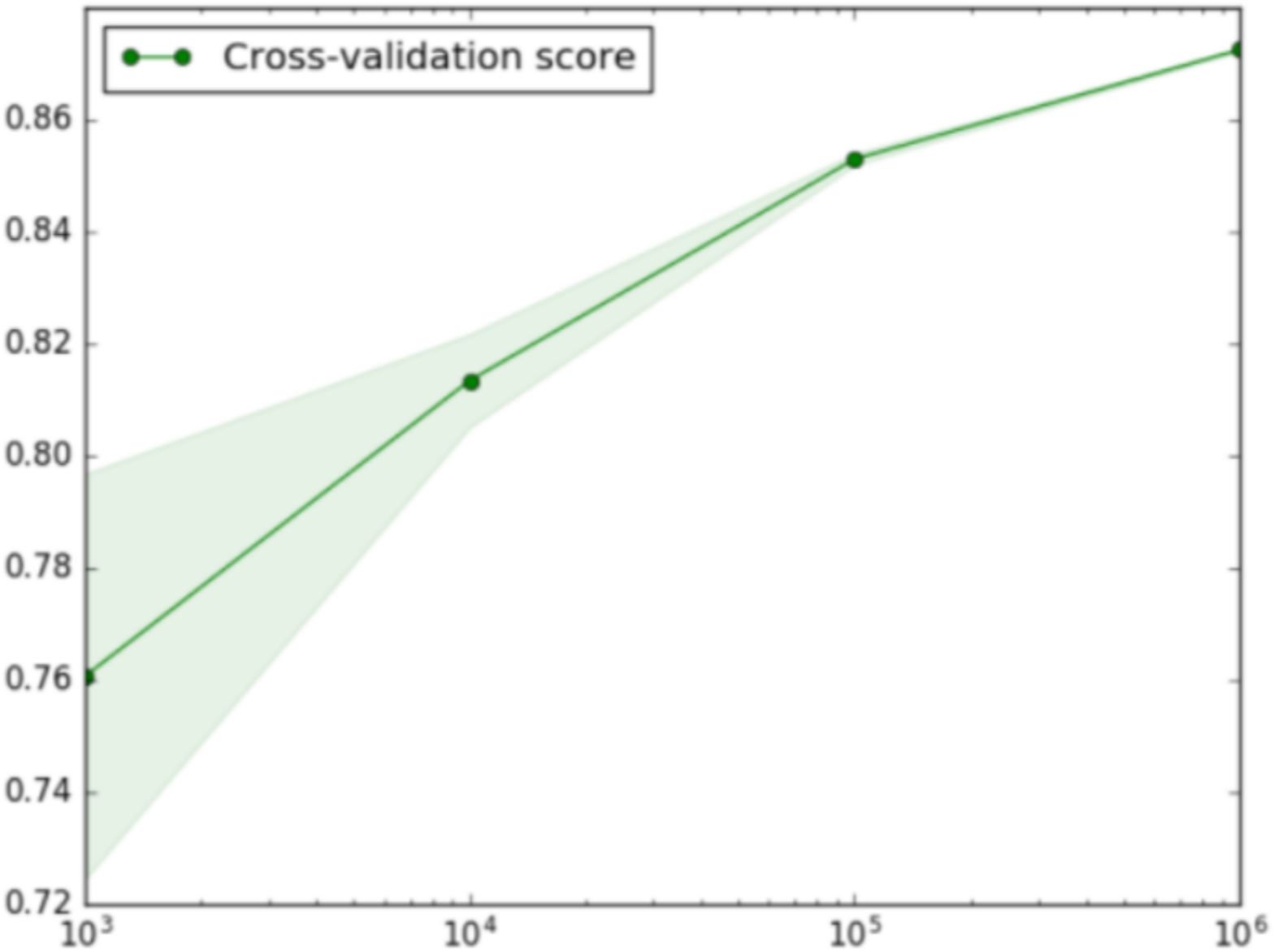}
  \caption{With bigrams}
  \label{fig4}
\end{subfigure}%
\begin{subfigure}{.5\textwidth}
  \centering
  \includegraphics[width=.85\linewidth]{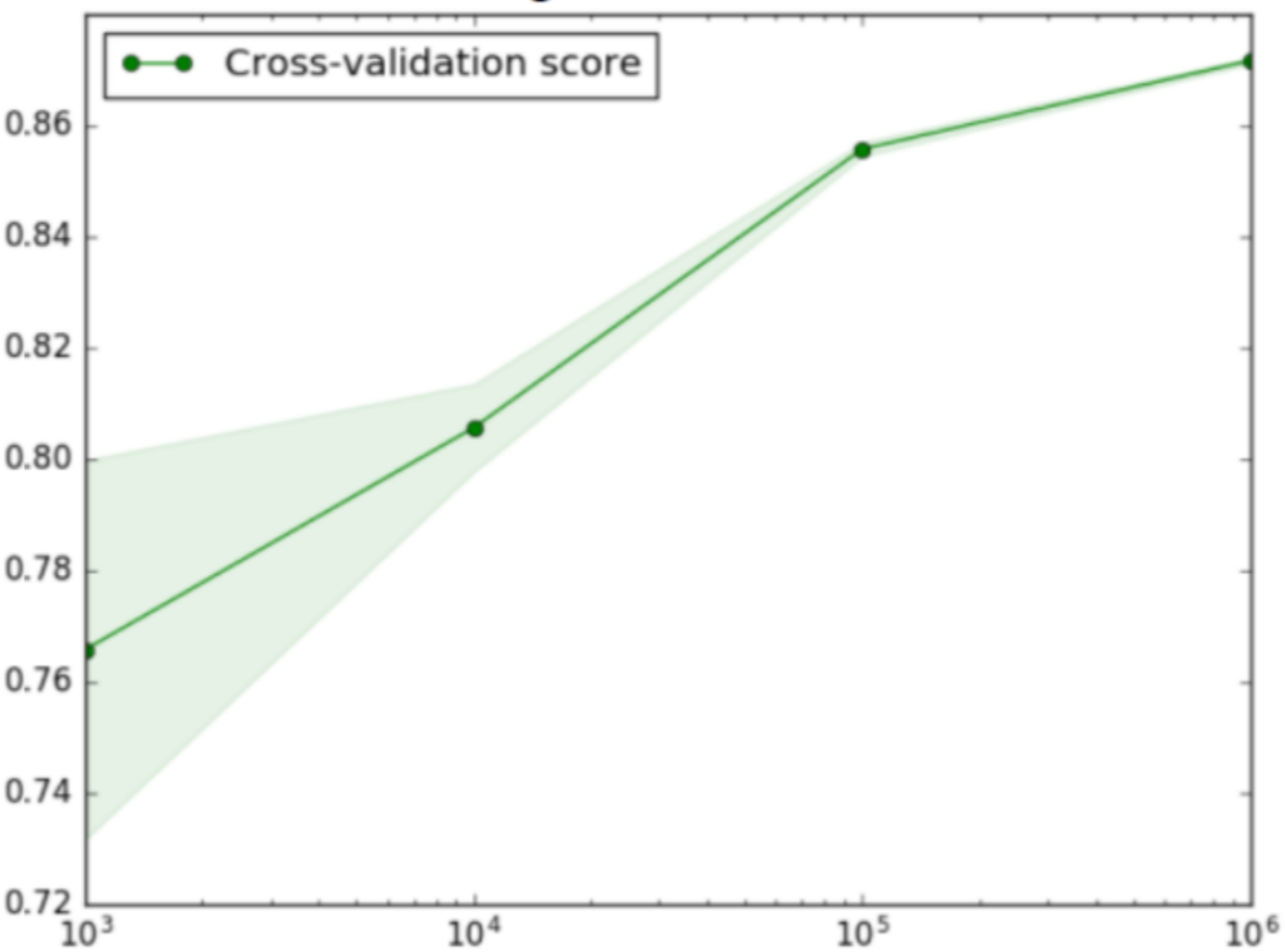}
  \caption{Without bigrams}
  \label{fig5}
\end{subfigure}
\vspace{-16pt}
\caption{Learning curves with 5-fold cross-validation on the FastText classifier using as input pre-trained word-embeddings learned (\textbf{a}) 
with bigrams and (\textbf{b}) without bigrams. The green region represents the standard deviation of the model.}
\label{fig3} 
\end{figure}

We also tested the predictive performance of this model when using or not using pregenerated bigrams, but there was no significant impact on the results, as shown in Figure \ref{fig3}.

The $F_1$ achieved when using $10^6$ samples ($800$K for training) was $0.87$ (see Table \ref{table_results}). The Cohen's Kappa coefficient of agreement between the predicted and the true labels in the validation set was $\kappa=0.75$, which is regarded as a substantial agreement. The results did not improve significantly when using the full dataset of 4.1 $\times$ $10^6$ samples. In this case, the $F_1$ achieved was 0.88 with a Cohen's Kappa coefficient  $\kappa=0.76$.

\begin{table}
\centering
\caption{FastText classifier hyper-parameters. {Optimization was performed using random search \cite{bergstra2012random}.}} 
\begin{footnotesize}
\begin{tabular}{ll}
\hline
\textbf{Hyper-parameter} & \textbf{Value} \\
\hline
Learning rate                         & 0.1         \\
Size of word vectors                  & 100           \\
Size of the context window            & 5             \\
Number of epochs                      & 100             \\
Minimum number of word occurences        & 1             \\
Number of negatives sampled              & 5             \\
Loss function                         & Softmax  \\
Minimum length of char n-gram             &  0            \\
Maximum length of char n-gram             &  0            \\
Maximum length of word n-gram             &  1            \\
Sampling threshold                    & $10^{-4}$       \\
Rate of updates for the learning rate & 100 \\ 
Use of pre-trained word vectors for supervised learning &Yes \\
\hline
\end{tabular}
\end{footnotesize}
\label{tab:w2v-hyper-classifier}
\end{table}
\unskip

The learning curve (Figure \ref{fig3}) shows the evolution of the $F_1$ during training when the number of training samples was increased from 800 to 800K. The curve converged with the score obtained in the training sample to a maximum of 0.88 when using the full dataset, as shown in Table \ref{table_results}. The validation score converges with the training score, and  the estimator does not benefit much from more training data, denoting a bias error. 
It has been reported that the phenomenon of not being able to increase the performance with additional data can be overcome with the use of deep learning models applied to complex problems, in contrast to a fast but thin architecture such as FastText (as will be proved later when using CNNs). On the contrary, the model did not suffer from a variance error. Cross-validation was used to assess how well the results of the model generalized to unseen datasets and obtained robust average validation results with a decreasing standard deviation over the $k$ folds (Figure \ref{fig:learningcurves}a).  




\subsubsection{CNN Classifier Results}

The hyper-parameters used to train the CNN model are shown in Table \ref{tab:cnn-hyper-classifier}. The results obtained when using both Gensim and FastText generated embeddings were studied, and the $F_1$ obtained was similar when using or not using pre-trained word-embeddings. Only the number of dimensions and epochs had a great impact on the performance and the computational cost of the model. 

\begin{table}
\centering
\caption{CNN classifier hyper-parameters.}
\begin{footnotesize}
\begin{tabular}{ll}
\hline
\textbf{Hyper-parameter} & \textbf{Value} \\
\hline
Batch size & 128 \\
Learning rate                & 0.001        \\
Size of word vectors                  & 100          \\
Number of epochs                      & 10             \\
Max number of words  	     & 20,000 \\
Max sequence length          &     1000        \\
Loss function                &  Categorical cross-entropy \\
Optimizer                   & RMSProp \\
RMSProp rho  & 0.9 \\ 
epsilon & $10^{-8}$ \\
decay & 0 \\
\hline
\end{tabular}
\end{footnotesize}
\label{tab:cnn-hyper-classifier}
\end{table}


With regard to the batch size, sizes of 1, 10, 64, 128, and 512 were investigated and, as expected, the higher the value, the greater the computational efficiency. The noisiness of the gradient estimate was reduced in batch sizes by using higher values. This can be explained by the fact that updating by one single sample is noisy when the sample is not a good representation of all the data. We should consider a batch with a size that is representative of the whole dataset. For values higher than 128, the~predictive performance deteriorated in earlier epochs during training and, therefore, we chose~a~value of 128. In fact, it has been reported that the loss function landscape of deep neural networks is such that large-batch methods are almost invariably attracted to regions with sharp minima \cite{Keskar_2016_OnLarge} and that, unlike~small-batch methods, they are unable to escape the basins of these minimizers. When using a~larger batch, there is consequently a significant degradation in the quality of the model, as measured by its ability to generalize.


The CNN learning curve (Figure \ref{fig:learningcurves}b) 
shows that the network is capable of generalizing well and that the model is robust. Unlike that which occurs with the FastText classifier, no overfitting is produced when the dataset is small. 

Nonetheless, it also had a bias error, but, in this case, the model achieved higher scores for both the training and the validation sets, converging to a maximum $F_1=0.91$, beyond which adding more data does not appear to be beneficial. 

{One additional difference with the FastText learning curve is that the CNN model needs more data to learn, in comparison with FastText. This is reflected by the fact that the CNN model was underfitting and not properly learning for a sample size $10^3$ with a validation score of only 0.72, while~for FastText and a sample size $10^3$, the model was clearly overfitting with a training score close to 1. }




The model of the whole dataset, using 3,257,179 training examples, bigrams, and pre-trained word-embeddings, eventually yielded an accuracy of 0.91 for the validation set comprising 814,295 samples. The coefficient of agreement between the predicted and the true labels in the validation set was $\kappa=0.83$ (see Table \ref{table_results}), which is regarded as an almost perfect agreement and implies that the model is reliable. 


\subsection{SVM Classifier Results}
{
The learning curve (Figure \ref{fig:learningcurves}c) shows the evolution of the  $F_1$ during training when the number of training samples was increased from 800 to 800K. The curve converged with the score obtained in the training sample to a maximum of 0.79 when using the full dataset, as shown in Table \ref{table_results}. The validation score converges with the training score, and the estimator does not benefit much from more training data, denoting a bias error.}


{The $F_1$ achieved when using $10^6$ samples ($800$K for training) was $0.79$ (see Table \ref{table_results}). The Cohen's Kappa coefficient of agreement between the predicted and the true labels in the validation set was $\kappa=0.57$. The results did not improve when using the full dataset of 4.1 $\times$ $10^6$ samples. In this case, the $F_1$ achieved was 0.79 with a Cohen's Kappa coefficient  $\kappa=0.58$.}

\subsection{kNN Classifier Results}
{The learning curve (Figure \ref{fig:learningcurves}d) shows the evolution of the  $F_1$ during training when the number of training samples was increased from 800 to 800K. The validation curve with a maximum of 0.92 still did not reach the training score obtained in the 800K training sample and further reached 0.93 in the full dataset, as shown in Table \ref{table_results}. This estimator benefited the most, compared with the other models, from more training data. Nonetheless, as expected, the computational cost on prediction time was expensive, and using $10^6$ samples was equivalent to 16 core-hours of CPU.}



{
The $F_1$ achieved when using $10^6$ samples ($800$K for training) was $0.92$ (see Table \ref{table_results}). The Cohen's Kappa coefficient of agreement between the predicted and the true labels in the validation set was $\kappa=0.83$, which is regarded as an almost perfect agreement. The results did not improve significantly when using the full dataset of 4.1 $\times$ $10^6$ samples. In this case, the $F_1$ achieved was 0.93 with a Cohen's Kappa coefficient  $\kappa=0.84$.
}

\subsubsection{Evaluation Using a Clinical Practice Simulation}

Finally, in order to assess the potential of the proposed approach as a clinical decision support system, we checked its performance using a clinical practice simulation. The two final models were, therefore, further tested with unseen inputs consisting of a small set (50 samples) of short clinical statements that would be used in routine clinical practice. Although the test size is too small to be able to draw meaningful conclusions, the models yielded very promising results with an accuracy of 0.88 and $\kappa=0.76$. This favors the hypothesis that it would be possible to generalize such a model to a different source of data (i.e., routine clinical practice notes) beyond clinical trial protocol eligibility criteria texts, which was the source used to build and validate it.


Some examples of correctly classified statements that would require an expert knowledge of oncology to judge whether or not they are cases being studied in available clinical trials (Yes/No) are shown below.
\newpage

\begin{itemize}
\item[] Lapatinib to treat breast cancer with brain metastasis $\rightarrow$  Yes;
\vspace{-0.3cm}
\item[] Pertuzumab to treat breast cancer with brain metastasis $\rightarrow$  No;
\vspace{-0.3cm}
\item[] CAR to treat lymphoma $\rightarrow$  Yes;
\vspace{-0.3cm}
\item[] TCR to treat breast cancer $\rightarrow$  No.
\end{itemize}

The performance achieved with the CNN classifier fits expectations with an $F_1 = 0.91$ and an~almost a perfect agreement, outperforming the FastText results. We can, therefore, conclude that it is possible to address the problem of predicting whether or not short clinical statements extracted from eligibility criteria are considered eligible in the available corpus of cancer clinical trials.

\subsection{Word-Embeddings}

The word-embeddings are an interesting part of this work. Adding pre-trained embeddings to the classifiers did not alter the classification results. However, the embeddings were, in themselves, sufficiently interesting to be qualitatively assessed and discussed using word space visualizations.

\subsubsection{t-SNE (t-Distributed Stochastic Neighbor Embedding) Representation of a Subset of Words}
\label{sec:tsnesubset}

{The word-embeddings obtained with FastText, in which each word is represented in a 100-dimensional space, can be used as a basis on which to visualize a subset of these words in a reduced space. We use t-Distributed Stochastic Neighbor Embedding (t-SNE \cite{VanDerMaaten_2008_VisualizingT-SNE}) for this purpose, which is a dimensionality reduction method that is particularly well suited to the visualization of high-dimensional datasets. The~objective of this algorithm is to compute the probability distribution of pairs of high-dimensional samples in such a way that similar prototypes will have a high probability of being clustered together. The algorithm subsequently projects these probabilities into the low-dimensional space and optimizes the distance with respect to the sample's location in that space. }

\begin{figure}
\centering
\includegraphics[width=\linewidth]{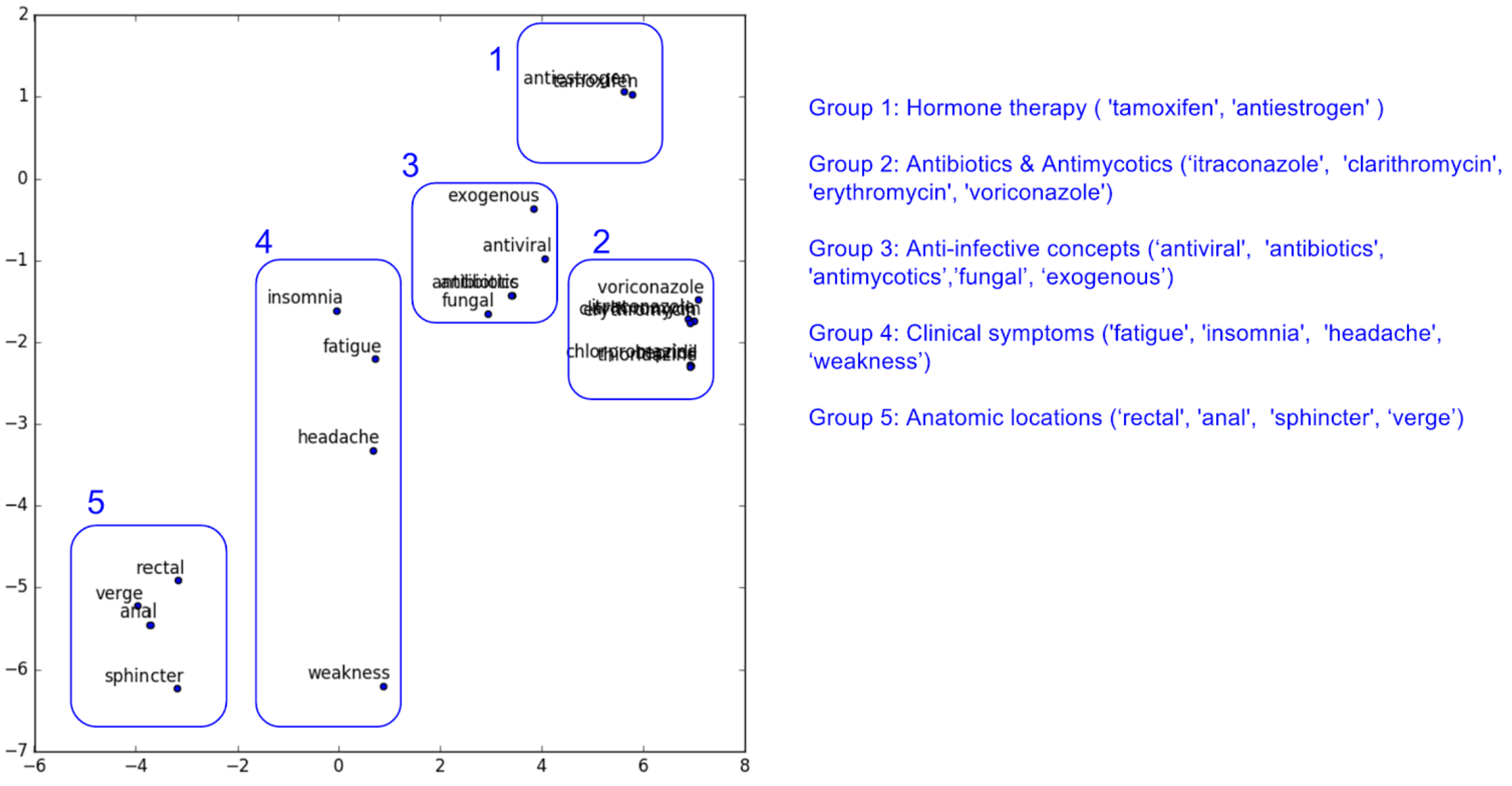}
\caption{Word-embeddings projected into a reduced space with t-Distributed Stochastic Neighbor Embedding (t-SNE).} 
\label{fig6}
\end{figure}

We defined those words from the complete corpus that we wished to analyze (as it is not possible to visualize all 26,893 words), and obtained the vectors of these words. The t-SNE representation in Figure~\ref{fig6} shows two aspects: on the one hand, the words are grouped by semantic similarities, and on the other, the clusters seem to follow a spatial distribution in different regions in a diagonal direction from intrinsic/internal to extrinsic/external concepts with respect to the human body: [G5]~body organs $\rightarrow$ [G4]~body symptoms $\rightarrow$ [G3]~infections, cancer and other diseases $\rightarrow$ [G1,G2]~treatments.


\subsubsection{Interactive Visualization of the Whole Set of Words}
\label{sec:interactivevis}

TensorBoard from TensorFlow \cite{Abadi_2016_TensorFlow} provides a built-in visualizer, called the Embedding Projector, for the interactive visualization and analysis of high-dimensional data. The Word2Vec embeddings obtained with Gensim were converted into Tensorflow 2D tensor and metadata formats for embedding~visualization. 

\begin{figure}
\centering
\includegraphics[width=\linewidth]{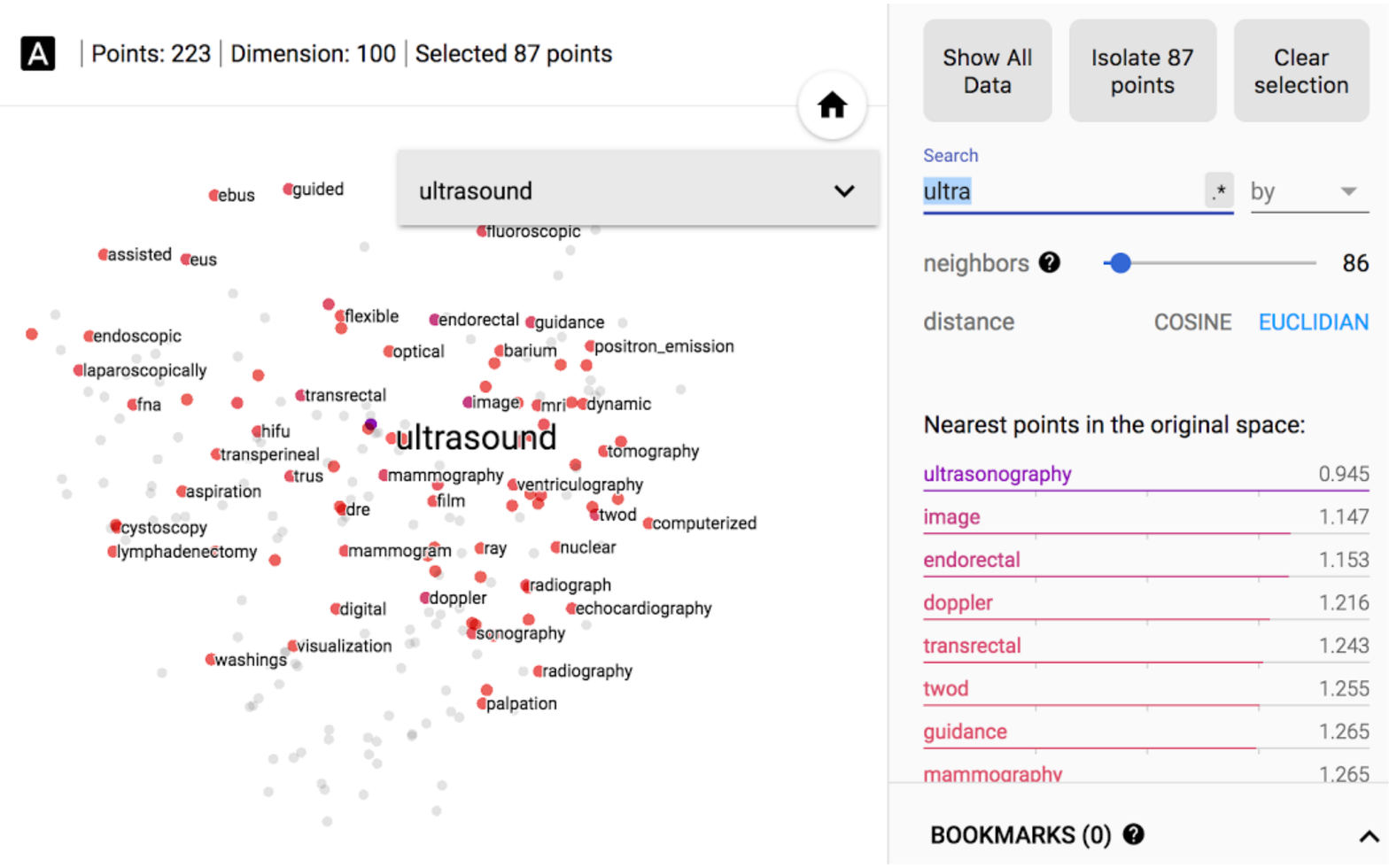}
\caption{Search for ``Ultrasound'' on the Tensorboard Embedding Projector.}
\label{fig7}
\end{figure}

Figure \ref{fig7} shows an example of these results when using the word ``ultrasound'' as a query. We~can appreciate that the 87 nearest points to ultrasound were all related to explorations, and mainly medical imaging. The nearest neighbor distances are also consistent when using other concepts. For example, Table \ref{tab:tamoxifen} shows that the model successfully extracted hormonal therapies from breast cancer as the t-SNE nearest neighbors to ``Tamoxifen''.

\unskip

\begin{table}
\centering
\caption{Nearest neighbors of ``Tamoxifen'' using Euclidean distance on the embedding t-SNE space. All of them are hormonal therapies.}
\label{tab:tamoxifen}
\begin{footnotesize}
\begin{tabular}{ll}
\hline
\textbf{Word} & \textbf{Distance }\\
\hline
Raloxifene & 0.569 \\
Letrozole & 0.635 \\
Anastrozole & 0.656 \\
Fulvestrant & 0.682 \\
Arimidex & 0.697 \\
Antiandrogens & 0.699 \\
Exemestane & 0.715 \\
Aromatase & 0.751 \\
Antiestrogens & 0.752 \\ 
Toremifene & 0.758 \\
Serm & 0.760 \\
Estrogens & 0.769 \\
Agonists & 0.773 \\
\hline\hline
\end{tabular}
\end{footnotesize}
\end{table}



\subsubsection{Word Clusters}
\label{sec:clusters}

We also used the resulting word vectors to generate word clusters fitting a $k$-means model \cite{kmeans}. The number of clusters were estimated by applying a reduction factor of $0.1$ to the total number of words to be read (maximum 10,000). The implementation and resulting clusters can be found at \url{https://github.com/auriml/capstone}. Upon sampling 20 clusters at random, a total of 16 were judged to be relevant as to whether their words were syntactically or semantically related. Some examples are shown in Table \ref{tab:clusteredTerms}.

\begin{table}
\centering
\caption{Samples of clustered words.}
\label{tab:clusteredTerms}
\begin{footnotesize}
\begin{tabularx}{\textwidth}{X|X|X|X}
mri, scan, imaging, radiographic, magnetic, resonance, scans, abdomen, radiological, mr, radiologic, image, technique, images, perfusion, sectional, weighted, spectroscopy, mris, dce, imaged, lp, neuroimaging, volumetric, mrs, multiparametric, mrsi, imagery 
& 
pelvis, skull, bones, skeleton, femur, ribs, sacrum, sternum, sacral, lfour, rib, humerus & 
pulmonary, respiratory, obstructive, asthma, copd, restrictive, emphysema, bronchiectasis, bronchodilator, bronchitis, bronchospasm, pneumothorax, ssc, bronchopulmonary, cor, expired, onel, congestion, airflow & 
abuse, alcohol, substance, dependence, alcoholism, addiction, dependency, illicit, recreational, user, illegal, misuse, abusers
\end{tabularx}
\end{footnotesize}
\end{table}

Note that medical abbreviations, such as \textit{lfour} (L4), \textit{mri} (Magnetic Resonance Imaging), or \textit{copd} (Chronic Obstructive Pulmonary Disease) were correctly clustered.

\subsubsection{Word Analogies}
\label{sec:wordanalogies}

The word vectors generated were also useful for accurately resolving analogy problems, such as  ``Tamoxifen is used to treat breast cancer as $X$ is used to treat prostate cancer?''. To find the top-N most similar words, we used the multiplicative combination “3CosMul” objective proposed by Levy \cite{levy2014linguistic}: 


\begin{itemize}
\begin{small}
\item[] \begin{texttt}
[`tamoxifen'  $-$ `breast + `prostate'] $\approx$
[(`enzalutamide', 0.998), (`antiandrogens', 0.972), (`abiraterone', 0.952), (`finasteride', 0.950), (`zoladex', 0.946), (`adt', 0.933), (`dutasteride', 0.927), (`acetate', 0.923), (`flutamide', 0.916), (`leuprolide', 0.910)]
\end{texttt}
\end{small}
\end{itemize}

These are, in fact, very precise results, because all these terms belong to the hormone-therapy family of drugs which are specifically used to treat prostatic cancer, and are the equivalents of tamoxifen (hormone-therapy) for breast cancer. In other words, the model learned the abstract concept ``hormone-therapy''  as a family of drugs and was able to apply it distinctively depending on the tumor type. 

\section{Conclusions}
\label{sec:conclusions}

In this work, we have trained, validated, and compared {various classifiers (FastText and a CNN with pre-trained word-embeddings, kNN, and SVM)} on a corpus of cancer clinical trial protocols (\url{www.clinicaltrials.gov}). 
The models classify short free-text sentences describing clinical information (medical history, concomitant medication, type and features of tumor, such as molecular profile, cancer therapy, etc.) as eligible or not eligible criteria for volunteering in these trials.
{
SVM yielded the lowest accuracy results, and kNN obtained top accuracy performance similar to the CNN model, but it had the lowest computational performance. Particularly, the high accuracy achieved with kNN is the immediate consequence of using as input a highly efficient clinical statement representation which is based on averaged pre-trained word-embeddings. A possible reason for this is that the kNN accuracy relies almost exclusively on using a highly efficient vector representation as the input data and on the dataset size. Being a non-parametric method, it is often successful---as in this case---in classification situations where the decision boundary is very irregular. Nonetheless, in spite of its high accuracy and the minimal training phase, we favor the use of deep learning architectures for classification (such~as CNN) over a kNN model because of its lower computational cost during prediction time. In fact, classifying a given observation requires a rundown of the whole dataset being too computationally expensive for large dataset as in this work.}

{All} models were evaluated using a 5-fold cross-validation on incremental sample sizes (1K, 10K, 100K, 1M, samples) and on the largest available balanced set (using undersampling) with 4.01 million labeled samples from a total of 6 million. Overall, the models proved robust and had the ability to generalize. The best performance {was achieved with kNN using a balanced sampling of the whole dataset. The results fit expectations, with an $F_1 = 0.93$ and an agreement of $\kappa=0.84$. The fact that the CNN model outperformed FastText may be explained by its greater depth, but more efforts should be made to experiment with alternative CNN topologies. }

This CNN model was also evaluated on an independent clinical data source, thus paving the way toward its potential use---taking into account pending improvements---in a clinical support system for oncologists when employing their clinical notes.

During the experiments, the word-embedding models achieved high-quality clusters, in addition to demonstrating their capacity for semantic reasoning, since they were able to identify the equivalent treatments for a type of tumor by means of an analogy with the drugs used to treat other tumors. These interesting reasoning qualities merit study in a future work using this dataset. 

The evaluation results show that clinical trial protocols related to cancer, which are freely available, can be meaningfully exploited by applying representation learning, including deep learning techniques, thus opening up the potential to explore more ambitious goals by making the additional efforts required to build the appropriate dataset.




{Our most immediate future work is to use a larger sample test of short clinical text from medical records for real simulation and include the effectiveness of CT interventions in the model, thus~enabling us to not only predict whether or not a patient’s case has been studied, but also whether the proposed treatment is expected to be effective based on the results of completed clinical trials for each indication. The problem would be a multilabel classification task, where the classes would be ``effective” vs. ``non-effective” and ``studied” vs. ``non-studied”, and both could be either true or false. This would allow us to classify from four types of cases: effective and studied, potentially effective but not studied, not effective and studied, and potentially not effective and not studied. The main effort in this case lies in the dataset building, which entails including the obtained efficacy results for each study. As only a subset of CTs (5754 samples, 11\%) have the results reported on \url{clinicaltrials.gov}, it means that, for this goal, it would be necessary to augment data from other sources, such as PubMed \cite{pubmed}.
Following~prior effort, a new model could be built to output potential cancer treatments that could be considered for a~particular patient case based on the efficacy results of completed clinical trials.}

\vspace{6pt}




\section*{Acknowledgment}
This work was supported by Medbravo, the Pattern Recognition and Artificial Intelligence Group (GRFIA) and the University Institute for Computing Research (IUII) from the University of Alicante. 


\section*{References}

%
\reftitle{References}

\end{document}